\def\year{2022}\relax
\documentclass[letterpaper]{article} 
\usepackage{aaai22}  
\usepackage{times}  
\usepackage{helvet}  
\usepackage{courier}  
\usepackage[hyphens]{url}  
\usepackage{graphicx} 
\usepackage{natbib}  
\usepackage{caption} 
\DeclareCaptionStyle{ruled}{labelfont=normalfont,labelsep=colon,strut=off} 
\usepackage{algorithm}
\usepackage{algorithmic}
\usepackage{booktabs}
\usepackage{fancyhdr}
\usepackage{amssymb}
\usepackage{amsmath}

\usepackage{newfloat}
\usepackage{listings}
\floatstyle{ruled}
\newfloat{listing}{tb}{lst}{}
\floatname{listing}{Listing}
\title{Spatial-Temporal Frequency Forgery Clue for Video Forgery Detection in VIS and NIR Scenario}
\author{Yukai Wang\textsuperscript{\rm 1}, Chunlei Peng\textsuperscript{\rm 1}, 
        Decheng Liu\textsuperscript{\rm 1}, 
        Nannan Wang\textsuperscript{\rm 1}, 
        and Xinbo~Gao\textsuperscript{\rm 2}}

\affiliations{
    \textsuperscript{\rm 1}Xidian University\\
    \textsuperscript{\rm 2}Chongqing University of Posts and Telecommunications\\

}

\usepackage{bibentry}

\begin{document}

\maketitle

\begin{abstract}
    In recent years, with the rapid development of face editing and generation, more and more fake videos are circulating on social media, which has caused extreme public concerns. Existing face forgery detection methods based on frequency domain find that the GAN forged images have obvious grid-like visual artifacts in the frequency spectrum compared to the real images. But for synthesized videos, these methods only confine to single frame and pay little attention to the most discriminative part and temporal frequency clue among different frames. To take full advantage of the rich information in video sequences, this paper performs video forgery detection on both spatial and temporal frequency domains and proposes a Discrete Cosine Transform-based Forgery Clue Augmentation Network (FCAN-DCT) to achieve a more comprehensive spatial-temporal feature representation. FCAN-DCT consists of a backbone network and two branches: Compact Feature Extraction (CFE) module  and Frequency Temporal Attention (FTA) module. We conduct thorough experimental assessments on two visible light (VIS) based datasets WildDeepfake and Celeb-DF (v2), and our self-built video forgery dataset DeepfakeNIR, which is the first video forgery dataset on near-infrared modality. The experimental results demonstrate the effectiveness of our method on detecting forgery videos in both VIS and NIR scenarios.
 
\end{abstract}

\section{Introduction}
In recent years, with the widespread application of Deep Neural Network (DNN) and Generative Adversarial Networks (GAN) technologies, face generation, and editing technology have become more and more realistic and controllable. However, the abuse of this technology has caused widespread public concern. The current mainstream face forgery technologies include face replacement \cite{1-li2020advancing, 2-chen2020simswap, 3-li2021faceinpainter, 4-zhu2021one, 5-agarwal2021detecting}, expression replay \cite{6-xue2020realistic, 7-yao2020mesh, 8-burkov2020neural, 9-song2021everything}, attribute editing \cite{10-gao2021high, 11-afifi2021histogan, 12-xu2021facecontroller, 13-9316964}, and non-target face generation \cite{14-pidhorskyi2020adversarial, 15-park2020contrastive, 16-karras2020analyzing}. Face replacement refers to replacing the entire face of one target with the face of another. Face expression replay refers to synchronizing the facial expression of the original video into the target video to reproduce the expression. Face attribute editing usually modifies specific attribute features of face images such as skin color, hair color style and whether to wear glasses or not, \textit{etc}. Non-target face generation refers to the generation of people who do not exist in the real world, which can be accomplished with the help of GAN. With the rapid development of these technologies, the generation of fake images or videos has become more and more controllable. It is extremely hard to distinguish the real and fake videos on  social media only with  naked eyes. Thus, the malicious spread of these fake videos on the Internet has arose great challenges for society and citizens. For example, forged videos may affect national security, especially tampering with inflammatory videos of some leaders to spread through the Internet may affect the relationship between the two countries and even lead to a war. Fake videos deliberately forged by criminals are likely to have serious public opinion impacts on citizens. On the other hand, near-infrared (NIR) technology has been widely used in heterogeneous face recognition systems because of its advantage that it is robust to illumination. \cite{wang2020facial} pointed out that due to the difference in spectral components between NIR and VIS modalities, visible light and near-infrared (VIS-NIR) face recognition is still a challenging task and proposed a method of converting visible light into near-infrared modality to help the near-infrared face recognition system. Furthermore, the NIR-based face recognition systems are also threatened by face forgery attacks. For example, an attacker can inject a fake near-infrared face into the NIR-based access control system to escape identity authentication and thus avoid tracking; the NIR-based face unlocking technology used in the Xiaomi 8 phone was successfully cracked by simply using a black and white printer printout and adding shadows around the eyes and nose of the face to give it a more three-dimensional appearance. In addition, near-infrared is often used for night video surveillance because it is robust to illumination. However, if the NIR surveillance images collected in the night scene are forged to create rumors out of nothing and instigate the spread of public opinion, it will damage the reputation of citizens \cite{park2017detecting}. Therefore, research on forgery detection in NIR modality is also very important. Considering the wide application of authentication under the NIR scenario, the severe challenges of face forgery and detection in the NIR modality should be considered \cite{22-wang2022forgerynir}.

To address these challenges, a lot of research efforts have been dedicated to the field of face forgery detection. Approaches for face forgery detection can be broadly divided into two categories image/frame-based and video-based methods.  [44] utilize the inconsistency in estimating 3D head pose from faces for deepfake detection. [23] combined Representation learning (ReL) and Knowledge Distillation (KD) to enhance the generalization ability of forgery detection models against different generative techniques. \cite{17-yu2019attributing} first proposed the GAN fingerprint phenomenon, it found that images generated by different GAN models leave diverse features which are called fingerprint features. Inspired by the principle that synthesized images in the frequency domains are often more able to reflect the visual artifacts, some attempt to obtain information from frequency domains, such as Discrete Fourier Transform (DFT) and Discrete Cosine Transform (DCT). For example,  \cite{18-frank2020leveraging} analyzed the real and GAN-generated images and found that the images manipulated by GANs perform obvious grid-like traces in the frequency domain. It is proved that the Discrete Cosine Transform (DCT) spectrum is more linearly separable than the RGB domain. Wang \textit{et al.} \cite{19-wang2020cnn} proposed that the GAN artifact on the DCT spectrum is arisen by the operation of upsampling. Extensive experiments show that adding perturbation can remove the subtle artifact to a certain extent. \cite{20-dzanic2020fourier} showed that deep network-generated images share an observable, systematic shortcoming in replicating the attributes of these high-frequency modes, and modeled this clue to detect forgery images. However, \cite{21-chandrasegaran2021closer} believes that the high-frequency Fourier spectral attenuation difference is not an inherent defect of CNN-generated models, and these shortcomings can be erased by modifying the upsampling operation. Recently, more and more researchers believe that the image/frame-based method ignores the temporal inconsistencies between frames. Therefore, they directly extract video-level features for forgery detection. \cite{47-li2020sharp} proposed Sharp Multiple Instance Learning (S-MIL) extract spatio-temporal features to fully capture the intra-frame and inter-frame inconsistency. \cite{48-chugh2020not} proposed a deepfake video detection method by comparing the feature differences between audio and video sequences. \textit{However, existing methods did not take both the spatial frequency domain and the temporal clue into account which restricts the performance and robustness of video forgery detection.}

In this paper, we aim to propose a Discrete Cosine Transform-based Forgery Clue Augmentation Network (FCAN-DCT) for video forgery detection to achieve a more comprehensive representation of fake videos. FCAN-DCT consists of a backbone network and two branches: Compact Feature Extraction (CFE) module and Frequency Temporal Attention (FTA) module. The backbone network first extracts the feature map of the video sequences and then converts the feature map of the RGB space to the frequency domain through 2D-DCT and the converted feature maps can be considered as frequency spectrums. However, due to the discreteness of the DCT transform, the spectrum information extracted by the backbone network is not compact. Considering that the frequency components in the $h$ and $w$ axes on the DCT spectrum gradually increase from low to high, CFE adopts a block strategy and uses the maximum value in each block to represent the weight of each of the frequency domain components. Therefore, a more compact spatial frequency domain representation is obtained. FTA uses the channel attention mechanism on the frequency spectrum to obtain the temporal attention map of the entire frame sequences and makes a feature fusion with FTA to extract the spatial-temporal features among different frames. Therefore, FCAN-DCT learns a comprehensive and compact forgery feature representation for forgery videos in the frequency domain. We have conducted sufficient experiments on WildDeepfake \cite{30-zi2020wilddeepfake} and Celeb-DF (v2) \cite{31-li2020celeb} respectively, and the experimental results have outperformed the state-of-the-art methods. In addition, face forgery detection in multimodal scenarios is rarely explored except for \cite{22-wang2022forgerynir} which is a NIR image forgery dataset. Therefore, we further construct a NIR video forgery dataset, DeepfakeNIR, to evaluate the robustness and generalization ability of our method and promote the research on video forgery detection in multimodal scenarios. Moreover, in order to sinumate the real scenerao and increase the challenge of our dataset, we also conduct some perturbations such as Gaussian blur, Gaussian white noise and JPEG compression on DeepfakeNIR and the perturbed DeepfakeNIR are called DeepfakeNIR++. In summary, the main contributions are as follows:
\begin{itemize}
\item{We focus on the spatial-temporal frequency forgery clue for video forgery detection by paying attention to the discriminative features among frame sequences, which fills the gap of video forgery detection in the frequency domain.}
\item{We propose a Discrete Cosine Transform-based Forgery Clue Augmentation Network (FCAN-DCT). The CFE and FTA modules make full use of the information in different video frames to obtain compact and discriminative features respectively. These two branches are combined to achieve a more comprehensive spatial-temporal feature representation in the frequency domain.}
\item{We firstly perform video forgery detection experiments on both VIS and NIR scenarios, including two wildly used datasets WildDeepfake, Celeb-DF (v2), and our newly constructed DeepfakeNIR dataset, to evaluate the effectiveness and robustness of the proposed method.}
\item{We present a novel video forgery detection dataset (DeepfakeNIR) with diverse distributions in terms of posture, occlusion, and expression. There are a total of 3,816 NIR videos, where the ratio of real and fake NIR videos is 1: 1. The proposed dataset can help facilitate the research of video forgery detection in heterogeneous modalities.}
\end{itemize}

\begin{figure*}[t]
  \centering
  \includegraphics[width=\linewidth]{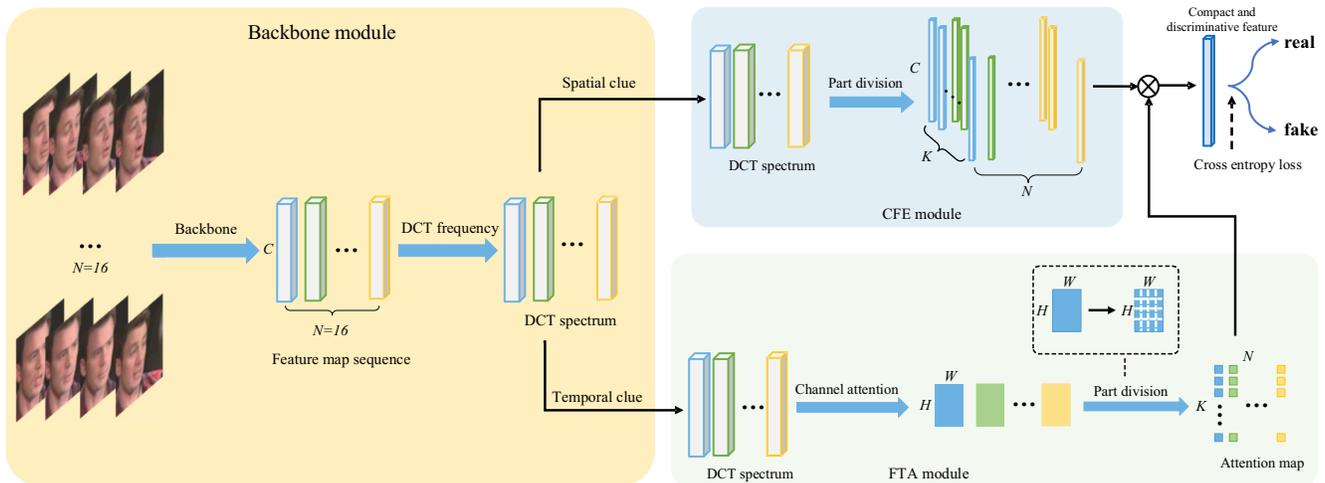}
  \caption{Framework of the Discrete Cosine Transform-based Forgery
Clue Augmentation Network (FCAN-DCT) for video forgery detection in VIS and NIR scenario.}
  
  \label{fig:framework}
\end{figure*}

\section{Related work}
In recent years, with the significant advancements made in the generation of deepfakes, tampered  facial images and videos have caused a significant amount of negative impact on social media, and also motivated methods to detect these forged videos. Existing methods define video forgery detection as a binary classification problem and develop image-level and video-level solutions. 
\subsection{Facial Image Forgery Detection}
Early face generation technologies often directly show obvious visual artifacts and inconsistencies in the facial area due to their uncontrollable characteristics. Many previous works employ face or head statistical inconsistencies for image forgery detection.  For example, the inconsistency in estimating 3D head pose from face images is introduced in \cite{51-yang2019exposing} to determine the authenticity of an image. Through the unremitting efforts of these works, facial image  forgery detection has achieved remarkable success in different aspects. \cite{46-yang2021learning} argues that the features extracted by the classification framework couple a lot of content-related information. It proposed a GAN Fingerprint Disentangling Network (GFD-Net) to disentangle the fingerprint feature  from GAN-generated images. \cite{45-kim2021fretal} perform domain adaptation tasks employing the Representation Learning (ReL) and Knowledge Distillation (KD) paradigms to counter with various deepfake generation techniques. However, with the development of DeepFake generation techniques, these image-based detectors may fail to capture temporal inconsistencies across multiple frames.  A large number of research efforts have been dedicated to exploring the video-level features.

\subsection{Facial Video Forgery Detection}
The early forgery generation technology is relatively limited, and it may show obvious temporal inconsistency between frames visually.  Physiological signal based on blink frequency which is not well represented in synthetic fake videos is explored in \cite{52-li2018ictu}.
 Agarwal \textit{et al.} \cite{53-agarwal2019protecting} found that the tampering with the face area in the video can cause the pattern of facial expressions and head movements inconsistent with the person's identity when speaking. In this way, soft-biometric models of individual national leaders were established, and these models were used to distinguish between real and fake videos. However, this approach is not universal. Recently, with the widespread application of Deep Neural Network (DNN) and Generative Adversarial Networks (GAN) technologies,
face generation and editing technology have become more and more realistic and controllable but it also brings more security issues to digital forensics. Aiming at the partial faces manipulation problem in DeepFake video detection, \cite{47-li2020sharp} proposed Sharp Multiple Instance Learning (S-MIL) in which  a spatial-temporal encoded instance is designed to fully capture the intra-frame and inter-frame inconsistency. \cite{48-chugh2020not} proposed a deepfake video detection method based on the dissimilarity between audio and video sequences. \cite{49-hosler2021deepfakes} proposed a novel method to justify the authenticity of videos by detecting the consistency of the emotions predicted by the speaker's face and voice. \cite{50-haliassos2021lips} detected forged videos based on high-level semantic irregularities in lip movements, which can be generalized to new forgery methods and various perturbation operations. \cite{zheng2021exploring} investigated the effective temporal cues for more robust video face forgery detection and proposed a general and flexible framework which combined fully temporal convolution network (FTCN) and Temporal Transformer to detect temporal incoherence explicitly.

\subsection{Frequency-based Forgery Detection}
 Considering that synthesized images in the frequency domain are often more able to reflect the visual artifacts, a lot of efforts have been dedicated to the field of forgery detection in the frequency domain, such as Discrete Fourier Transform (DFT) and Discrete Cosine Transform (DCT). \cite{20-dzanic2020fourier} showed that the decay rate of the high-frequency component in the Fourier spectrum is clearly distinguishable between real and deep network generated images. Using this clue, it proposed a novel model to detect synthesized images. \cite{21-chandrasegaran2021closer} argued that the high-frequency Fourier spectral attenuation difference is not an inherent feature of the CNN-based generative model, and this forgery clue can be erased by slightly modifying the upsampling operation, so the identification based on this feature is not robust. \cite{38-qian2020thinking} conducted DCT operation on the image, and then inversely transforms back to the spatial domain based on the low-frequency, mid-frequency and high-frequency information to obtain the spatial components of different frequency bands, which were finally sent to CNN to extract features. \cite{24-li2021frequency} argued that fixed filters and hand-crafted features are not sufficient to extract the forgery feature from the frequency domain. Therefore, single-center loss (SCL) was proposed to make the intra-class features of the real class more compact. All the aforementioned frequency-based methods confine to exploring the subtle spatial artifact but pay little attention to the temporal frequency clue among different frames. Our proposed FCAN-DCT not only explores the spatial frequency feature but also augments the temporal clue in the frequency domain with the attention mechanism. Thus, The final spatial-temporal feature learned by our method can achieve a more comprehensive representation.

\section{PROPOSED APPROACH}
Aiming at filling the gap of analyzing the entire video sequence instead of a single image in the frequency domain, we introduce the Discrete Cosine Transform-based Forgery Clue Augmentation Network (FCAN-DCT) for forgery detection. FCAN-DCT consists of a Compact Feature Extraction (CFE) module in section \ref{CFE} and a Frequency Temporal Attention (FTA) module in section \ref{FDA}. Fig. \ref{fig:framework} shows the framework of FCAN-DCT. The goal is to make full use of the rich information in the frame sequences which is not considered in the aforementioned methods, explore the spatial-temporal frequency clue among multiple frames, and finally achieve a more comprehensive feature representation in the frequency domain. 

\subsection{Preliminaries}
Since the tampering area in the forgery task is only limited to the face area, in order to prevent other areas from affecting the final detection result, we first use the Facenet \cite{29-schroff2015facenet} to extract the landmarks of the face and then extract the face area in each frame. Then, we uniformly sample $N$ cropped frames as the video input sequence of the model 
$I = \left\{ {{f_1},\;{f_2},\;{f_3},\; \ldots \;{f_N}} \right\}$. Each selected frame is input to the backbone network and finally outputs the feature map representation of each frame 
$M =\left\{ {{m_1},\;{m_2},\;{m_3} \ldots \;{m_N}} \right\}$. We transform these resulting feature maps into the frequency domain using the Discrete Cosine Transform (DCT). Much like the Discrete Fourier Transform (DFT), DCT represents a sequence of points in space as the sum of cosine functions at different frequencies. Given a pixel matrix in RGB space ${\text{f}} \in {\mathbb{R}^{N \times N}}$, the 2D-DCT is defined as:
\begin{tiny}
    \begin{equation}
    \label{equation-1}
    \begin{gathered}
    D\left( {u,v} \right) = c\left( u \right)c\left( v \right)\mathop \sum \limits_{i = 0}^N \mathop \sum \limits_{j = 0}^N f\left( {i,j} \right)\cos \left[ {\frac{{\left( {i + 0.5} \right)\pi }}{N}u} \right]\cos \left[ {\frac{{\left( {j + 0.5} \right)\pi }}{N}v} \right]
    \end{gathered}
    \end{equation}
\end{tiny}

\begin{small}
\begin{center}
\begin{equation}
c\left( u \right) = \left\{ {\begin{array}{*{20}{c}}
  {\sqrt {\frac{1}{N}} ,\;\;\;\;u = 0} \\ 
  {\sqrt {\frac{2}{N}} ,\;\;\;\;u \ne 0} 
\end{array}} \right.    
\end{equation}
\end{center}    
\end{small}

where $u$, $v$ are the coordinates of the DCT spectrum matrix 
$D\left( {u, v} \right) \in {\mathbb{R}^{N \times N}}$. Compared with DFT, DCT has better energy concentration in the frequency domain, and can directly crop those unimportant frequency domain regions and coefficients, so it is widely used in image compression tasks, such as jpeg compression. In recent years, with the rapid development of deep learning, the operation of DCT also attracted considerable interest in forgery detection tasks.

\begin{figure}[t]

  \centering
  \includegraphics[width=0.5\linewidth]{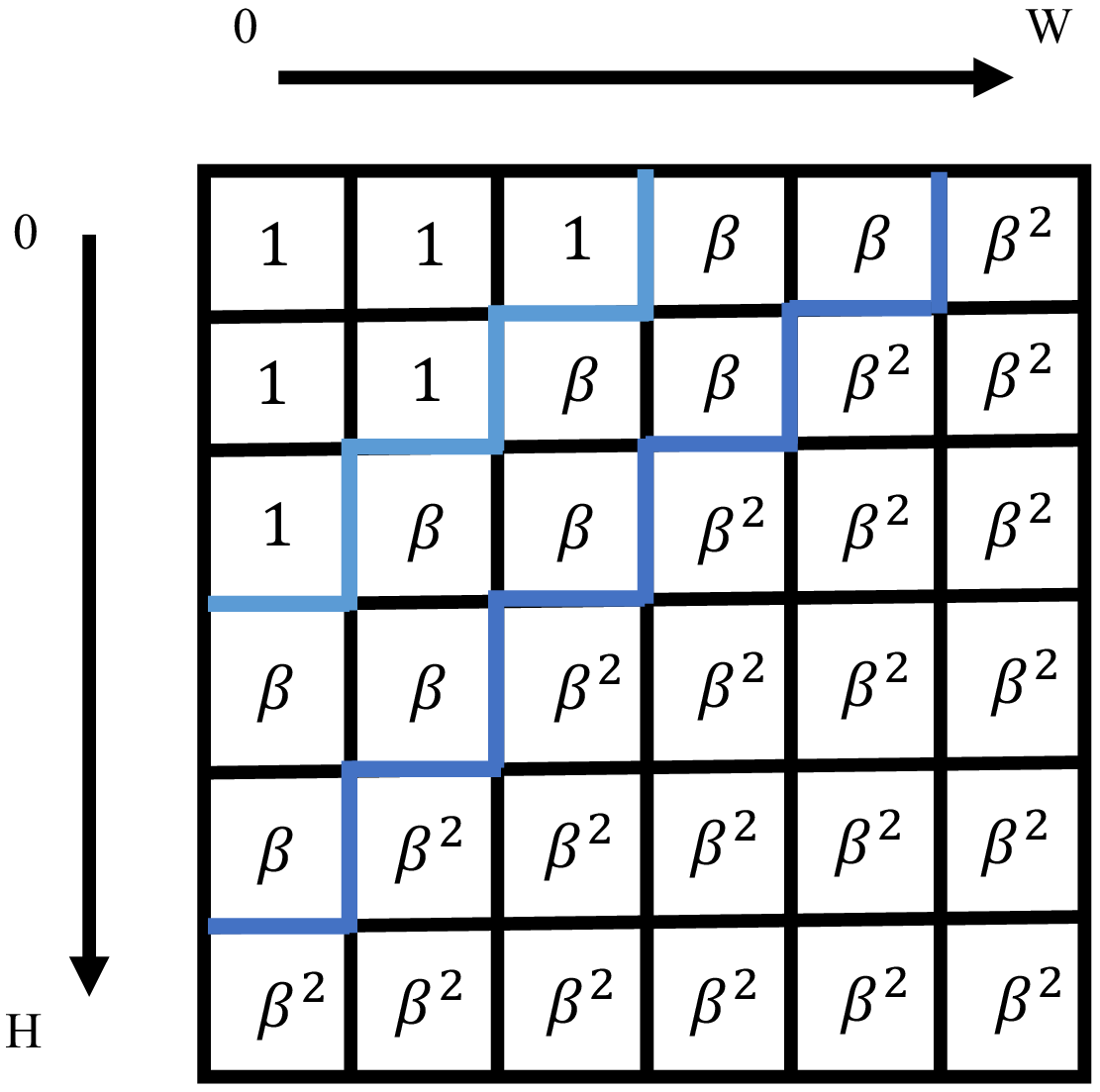}
  \caption{The weight matrix $W$ for the 2D-DCT spectrum. We learn discriminative features by enhancing the intensity of midium and high-frequency components in the frequency domain to amplify the difference between real and fake images.}
  
  \label{fig:weight_matrix}
\end{figure}

\subsection{Compact Feature Extraction Module}
\label{CFE}
We adopt Compact Feature Extraction (CFE) module to present a compact spatial feature representation.
CFE is mainly divided into two steps: 1) Use the 2D-DCT in equation (\ref{equation-1}) to transform the input pixel feature map $M = \left\{ {{m_1},\;{m_2},\;{m_3} \ldots \;{m_N}} \right\}$  to the DCT frequency domain $D = \left\{ {{d_1},\;{d_2},\;{d_3} \ldots \;{d_N}} \right\}$, while retaining more important information and discarding unimportant information. At the same time, artifacts that appear to be fake images are shown. 2) Compress the obtained frequency domain sequence to obtain a more compact feature representation. Let
${\text{M}} \in {{\mathbb{R}}^{N \times C \times H \times W}}$ represent the feature map sequence of the frame set extracted by the backbone network. According to equation (\ref{equation-1}) mentioned above, we convert the obtained 2D spatial pixel feature map to the frequency domain, and the obtained 2D spectrum is recorded as ${\text{D}} \in {{\mathbb{R}}^{N \times C \times H \times W}}$. Inspired by \cite{18-frank2020leveraging, 19-wang2020cnn, 20-dzanic2020fourier, 21-chandrasegaran2021closer, 24-li2021frequency}, which think the component present on the spectrum in midium and high-frequency bands is pretty inconsistent between real and synthesized images, we believe that we should make use of the differences in the midium and high-frequency components as much as possible and increase the degree of influence of these parts in the frequency spectrum. Therefore, a weight matrix $W$ is designed as Fig. \ref{fig:weight_matrix}. Here, $ \beta = \sqrt{2} $. For 2D DCT, the equation can be expressed as:


   


\begin{equation}
F\left( {u,\;v} \right) = {\beta ^\alpha }D\left( {u,\;v} \right),\;\;\;\;\left\{ {\begin{array}{*{20}{c}}
{\alpha  = 0,}&{0 \le u + v < \frac{H}{3}}\\
{\alpha  = 1,}&{\frac{H}{3} \le u + v \le \frac{{2H}}{3}}\\
{\alpha  = 2,}&{\frac{{2H}}{3} < u + v < H}
\end{array}} \right.
\end{equation} \\
where $u$, $v$ are the coordinates of the DCT spectrum, and $H$ is the height of the spectrum.

After multiplication with the coefficient matrix, the resulting spectrum sequence is $ {\text{F}} \in {{\mathbb{R}}^{N \times C \times H \times W}} $.
Considering the discrete nature of DCT transform, that is, some of the most valuable and effective information is often distributed in different parts, we adopt a block strategy and obtain the final compact spatial feature. First, the resulting DCT spectrum $ F(i) $ is divided into $K$ blocks, then the maximum value of each block is used as the expression of the block, and the compact feature map 
${{\text{F}}_{{\text{CFE}}}} \in {{\mathbb{R}}^{N \times C \times K}}$is obtained as the final representation.

\subsection{Frequency Temporal Attention Module}
\label{FDA}
To further explore the temporal frequency clue among multiple frames, we design a Frequency Temporal Attention (FTA) module. Similar to CFE, the input of the FTA module is also the DCT spectrum
${\text{F}} \in {{\mathbb{R}}^{N \times C \times H \times W}}$. The goal of this module is to construct an attention map ${\text{A}} \in {{\mathbb{R}}^{N \times H \times W}}$ based on the DCT frequency spectrum. First, we conduct the $L2$ normalization in the channel dimension. 

\begin{equation}
A\left( {n,h,w} \right) = \sqrt {\mathop \sum \limits_{i = 0}^{C - 1} {{\left( {{F_{n,i,h,w}}} \right)}^2}} 
\end{equation}
where $C$ is the number of channels. $A\left( {n,h,w} \right)$ represents the attention score of the $n^{th} $ frame spectral feature map at position $(h, w)$.
Then in order to control all the values in the attention matrix to be between 0-1, the results are normalized as:
\begin{equation}
A'\left( {n,h,w} \right) = \frac{{A\left( {n,h,w} \right)}}{{\mathop \sum \nolimits_{i = 0}^{H - 1} \mathop \sum \nolimits_{j = 0}^{W - 1} A\left( {n,i,j} \right)}}
\end{equation}

Following the same principle in CFE, we also divide the feature map matrix into $K$ blocks, and for each region, we use the sum of them as the attention score. The results are :
$$ A''\left( n \right) = \left\{ {A''\left( {n,1} \right),\;A''\left( {n,2} \right) \ldots \;A''\left( {n,K} \right)} \right\}$$ Finally, $L1$ norm based on ${A''} \in {{\mathbb{R}}^{N \times K}}$ is adopted to further explore the temporal clue between frame sequences. The 2D attention matrix is expressed as:
\begin{equation}
{A_{FTA}}\left( {n,k} \right) = \frac{{A''\left( {n,k} \right)}}{{\mathop \sum \nolimits_{i = 0}^{K - 1} A''\left( {n,i} \right)}}
\end{equation}

In the end, the spatial frequency forgery clue ${F_{CFE}} \in {{\mathbb{R}}^{N \times C \times K}}$ and the temporal frequency attention map ${A_{FTA}} \in {{\mathbb{R}}^{N \times K}}$ are merged together by applying the weights obtained with FTA module to the spatial feature in the  CFE module. The final spatial-temporal frequency forgery feature ${f_c} \in {{\mathbb{R}}^{C}} $ is constructed as: 
\begin{equation}
{f_c} = \mathop \sum \limits_{n = 0}^{N - 1} \mathop \sum \limits_{k = 0}^{K - 1} {F_{CFE}}\left( {n,c,k} \right) \times {A_{FTA}}\left( {n,k} \right)
\end{equation}


\section{EXPERIMENTS AND RESULTS}
In this section, we first introduce the benchmarks and the parameter settings in the experiments. Furthermore, we conduct sufficient experiments to demonstrate the effectiveness of our method FCAN-DCT. The results on the self-built near-infrared modality dataset DeepfakeNIR for face forgery detection also demonstrate that FCAN-DCT has good generalization across NIR and VIS modalities.

\subsection{Experimental Settings}
\begin{figure}[t]
  \centering
  \includegraphics[width=\linewidth]{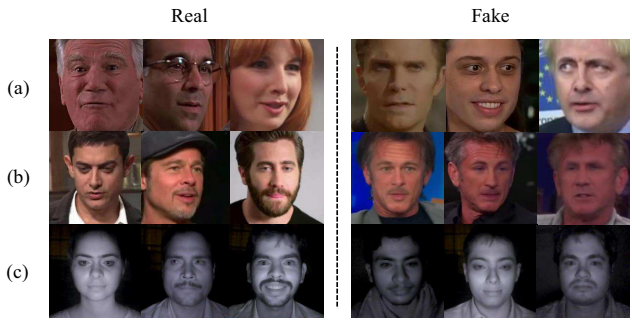}
  \caption{We conduct experiments on three datasets to verify the effectiveness of FCAN-DCT. The first two benchmarks are common datasets based on the visible light (VIS) modality: (a)WildDeepfake. (b) Celeb-DF (v2). The other is our proposed video dataset based on near-infrared (NIR) modality: (c) DeepfakeNIR.}
  
  \label{fig:datasets}
\end{figure}

\begin{figure}[t]
  \centering
  \includegraphics[width=\linewidth]{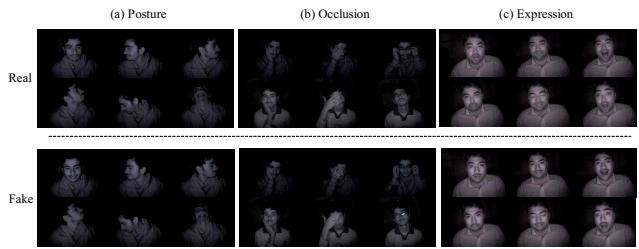}
  \caption{Example frames in our DeepfakeNIR with a diverse distribution in terms of posture, occlusion and expression.}

  \label{fig:FNIR-V}
\end{figure}

{\bfseries Datasets}: In this paper, we use four video-based datasets for forgery detection: WildDeepfake \cite{30-zi2020wilddeepfake}, Celeb-DF (v2) \cite{31-li2020celeb}, FaceForensics++ \cite{rossler2019faceforensics++} and our newly constructed DeepfakeNIR datasets. 
\begin{itemize}
\item{{\bfseries WildDeepfake}  contains 7,314 facial sequences extracted from 707 Deepfake videos. The videos are all collected from bilibili and YouTube, and their face-swapping videos are synthesized by various methods, so the detection is more difficult. Examples are shown in Fig. \ref{fig:datasets}.}
\item{{\bfseries Celeb-DF (v2)}  is a large-scale dataset proposed on the basis of v1, which contains 590 original real videos collected from YouTube and 5,639 corresponding fake videos with diverse distribution in terms of gender, age, and ethnic group. Examples are shown in  Fig. \ref{fig:datasets}. }
\item{{\bfseries FaceForensics++ (FF++) } is a large-scale benchmark dataset which contains 1,000 original videos from youtube and 4,000 fake videos generated by four typical manipulation methods: \textit{i.e.}, Deepfakes (DF), Face2Face (F2F), FaceSwap (FS) and NeuralTextures (NT). Each method generates 1,000 fake videos corresponding to the original video. There are three versions of FF++ in terms of compression level, \textit{i.e.}, raw, lightly compressed (HQ), and heavily compressed (LQ).}
\item{{\bfseries DeepfakeNIR}: In addition to the visible light (VIS) modality, we also experimented with our method in the near-infrared (NIR) modality. DeepfakeNIR contains 3,816 videos in total, where the ratio of real and fake videos is 1: 1, where the real videos are collected from CASIA NIR-VIS 2.0 \cite{li2013casia} and the fake videos are generated using a well-designed deepfacelab tool \cite{perov2020deepfacelab}. Furthermore, we add some perturbations such as JPEG compression, Change in color contrast, Gaussian blur, and Gaussian white noise,  \textit{etc}. on the NIR videos to mimic the real-world scenarios. The new dataset are called DeepfakeNIR++. The example video frames in our DeepfakeNIR dataset are shown in Fig. \ref{fig:FNIR-V}.}
\end{itemize}

{\bfseries Preprocessing}: We select 16 frames from each video, then for each frame, we extract facial key points and crop out the head region of the person through the Facenet\cite{29-schroff2015facenet}.  Finally, we normalize all the faces with ImageNet mean [0.485, 0.456, 0.406] and standard deviation [0.229, 0.224, 0.225], and resize them to a fixed size which is $256\times256$ for ResNet50 \cite{he2016deep} and $299\times299$ for Xception \cite{34-chollet2017xception}. 

{\bfseries Training}: In our experiments, we use ResNet50 \cite{he2016deep} or Xception \cite{34-chollet2017xception} as the final backbone of our approach. For retaining more information on the video sequence, we just remove the last average pooling and fully connected layer. We optimize the networks by Adam optimizer with the beta1 = 0.9 and beta2 = 0.999 . The start learning rate lr = 0.0001 and it drops by 10 every time the accuracy does not increase after 5 consecutive epochs. The evaluation metrics in our experiments are accuracy and AUC.

{\bfseries Selection of division number $ K $ and the number of frames $N$ in each video}: We conduct experiments on a total of four combinations of  $ K=2\times2 $, $ 4\times4 $ and frames=8, 16. The results are shown in Table \ref{tab:k_frames}. It can be seen that when the number of blocks is $K$ = $ 4\times4 $ and the number of frames selected is 16, the accuracy of the model reaches the highest of 86.35\%, and the AUC is also the second highest at 93.87\%. However, this does not mean that more blocks or more frame selections are effective for model training. When $K$ is fixed to $ 2\times2 $, as the frames increase from 8 to 16 frames, the accuracy of the model dropped from 85.11\% to 84.74\%, and the AUC also dropped from 93.87\% to 92.40\%. Therefore, in the selection process of $K$ and frames, these two parameters cannot be considered separately. When we fix $N$ to 8, there is only a slight increase in accuracy but when $N$ is 16, increasing $K$ from $ 2\times2 $ to $ 4\times4 $, the accuracy increases by more than 1.5\% from 84.74\% to 86.35\%. So we choose $K$ = $ 4\times4 $, $N$ = 16 as the final parameter selection result.

\begin{table}[t]
  \caption{Experimental results on varying parameter $K$ and $N$ in which $K$ is the number of division parts, $N$ is the selected number of frames in each video. The backbone is ResNet50. All the experiments are conducted on the WildDeepfake benchmark. The best results are bolded.}
  \begin{tabular}{cccc}
    \toprule
    $K (parts)$ & $N (frames)$ & Accuracy (\%) & AUC (\%)\\
    \midrule
     $2\times2$  & 8 & 85.11 & \textbf{93.87} \\
     $2\times2$   & 16 & 84.74 & 92.40 \\
     $4\times4$  & 8& 85.12 & 91.66\\
     $4\times4$ & 16& \textbf{86.35} & 93.74 \\
    \bottomrule
  \end{tabular}
  \label{tab:k_frames}
\end{table}

\begin{table}[t]
  \caption{Experimental results on WildDeepfake dataset, where the best result is marked in bold and the next best result is underlined.}
  \begin{tabular}{ccc}
    \toprule
    Method & Accuracy (\%) & AUC (\%)\\
    \midrule
    ResNetV2-50   & 63.99 & -\\
    Xception  & 79.99& 88.86\\
     MesoNet  & 73.95 & 83.21 \\
     Capsule  & 78.68& 86.31\\
     ADD-Xception  & 79.23& -\\
     Zhou \textit{et al.}  & 79.87& 87.82\\
     ${{\text{F}}^3}$-Net  & 80.66& 87.53\\
     DAM  & 83.32& -\\
     PEL  & 84.14& 91.62\\
     \hline
     FCAN-DCT + Xception & \underline{86.23} & \underline{93.57} \\
     FCAN-DCT + ResNet50 & \textbf{86.35} & \textbf{93.74}\\
    \bottomrule
  \end{tabular}
  \label{tab:test_in_wildDeepfake}
\end{table}

\subsection{Experiments of Video Forgery Detection in VIS Scenario}
In this section, we conduct experiments to compare our method with recent methods on WildDeepfake \cite{30-zi2020wilddeepfake} and Celeb-DF (v2) \cite{31-li2020celeb} datasets. The experimental results are shown in Table \ref{tab:test_in_wildDeepfake} and  Table \ref{tab:test_in_celeb-df}. Our method uses ResNet50 \cite{he2016deep} or Xception \cite{34-chollet2017xception} as the backbone network which is commonly applied in forgery detection tasks, respectively. It can be seen that our method achieves  state-of-the-art performance on both WildDeepfake and Celeb-DF (v2) datasets. The superior performance demonstrates the feasibility and effectiveness of our proposed method.

{\bfseries Evaluation on WildDeepfake}: We choose ResNet50 and Xception as the backbone respectively and conduct extensive experiments on the WildDeepfake dataset. We compare FCAN-DCT with several recent state-of-the-art methods for face forgery detection including \textbf{MesoNet} \cite{32-afchar2018mesonet}, \textbf{Xception} \cite{34-chollet2017xception}, \textbf{Capsule} \cite{35-nguyen2019capsule}, \textbf{ADD-Xception} \cite{36-khormali2021add}, \textbf{Zhou et al} \cite{37-zhou2021face}. \textbf{${{\text{F}}^3}$-Net} \cite{38-qian2020thinking}, \textbf{DAM} \cite{39-lin2021improved}, and \textbf{PEL} \cite{40-gu2021exploiting}.

The experimental results are shown in Table \ref{tab:test_in_wildDeepfake}. It can be seen that our method achieves the highest accuracy of 86.35\% and AUC of 93.74\%. Compared to the frame-level method, ResNetV2-50 \cite{33-he2016identity} only achieves 63.99\% accuracy, and our method improves the accuracy by 22.36\%. Our method also increases the accuracy and AUC by 6.24\% and 4.71\% respectively on the basis of Xception \cite{34-chollet2017xception}. The superiority demonstrates the spatial-temporal feature extracted by FCAN-DCT is efficient and effective. For the selection of the backbone network, when we select the more complex Xception network, the accuracy and AUC are slightly decreased by 0.12\% and 0.17\%. Meanwhile, several frequency-based approaches with competitive results should be noticed. ${{\text{F}}^3}$-Net \cite{38-qian2020thinking} found that fake images can show not only subtle forgery traces but compression errors in the frequency domain. PEL \cite{40-gu2021exploiting} extracted fine-grained information hidden in the frequency domain as auxiliary features, which were finally fused with the features extracted in the RGB space. The two aforementioned methods demonstrate the effectiveness of expressing features in the frequency domain. Compared with ${{\text{F}}^3}$-Net \cite{38-qian2020thinking} and PEL \cite{40-gu2021exploiting}, the accuracy of our proposed method increased by 5.69\% and 2.21\% and AUC increased by 6.21\% and 2.12\% respectively, The remarkable progress demonstrates the superiority of FCAN-DCT.

\begin{table}[t]
  \caption{Experimental results on Celeb-DF (v2) dataset, where the best result is marked in bold and the next best result is underlined.}
  \begin{tabular}{ccc}
    \toprule
    Method & Accuracy (\%) & AUC (\%)\\
    \midrule
     RCN  & 76.25 & 74.87 \\
     R2Plus1D   & 98.07 & 99.43\\
     I3D  & 92.28& 97.59\\
     MC3  & 97.49& 99.30\\
     R3D  & 98.26& 99.73\\
     \hline
     FCAN-DCT + Xception & \underline{98.63} & \underline{99.95} \\
     FCAN-DCT + ResNet50 & \textbf{99.80} & \textbf{99.99}\\
    \bottomrule
  \end{tabular}
  \label{tab:test_in_celeb-df}
\end{table}

\begin{table*}[t]
\centering
  \caption{Experimental results on DeepfakeNIR and DeepfakeNIR++ dataset. where the best result is marked in bold.}
  \begin{tabular}{ccccc}
    \toprule
    Method & Train & Test & Accuracy (\%) & AUC (\%)\\
    \midrule
    CNNDetection  & WildDeepfake& DeepfakeNIR & 71.00 & 91.48 \\
    GANFingerprint  & WildDeepfake& DeepfakeNIR & 53.14 & 51.70 \\
    FCAN-DCT + ResNet50 & WildDeepfake& DeepfakeNIR & 91.40 & 97.33\\
    FCAN-DCT + Xception & WildDeepfake& DeepfakeNIR & \textbf{96.19} & \textbf{99.29}\\
    
    \hline
    CNNDetection  & Celeb-DF (v2) & DeepfakeNIR & 69.32 & 91.99 \\
    GANFingerprint  & Celeb-DF (v2) & DeepfakeNIR & 51.72 & 58.79 \\
    FCAN-DCT + ResNet50 & Celeb-DF (v2) & DeepfakeNIR & 91.05 & 99.51\\
    FCAN-DCT + Xception & Celeb-DF (v2) & DeepfakeNIR & \textbf{98.32} & \textbf{99.90}\\
    
    \hline
    CNNDetection  & DeepfakeNIR & DeepfakeNIR & 100 & 100 \\
    GANFingerprint  & DeepfakeNIR & DeepfakeNIR & 99.11 & 99.97 \\
    FCAN-DCT + ResNet50 & DeepfakeNIR & DeepfakeNIR &100 & 100 \\
    FCAN-DCT + Xception & DeepfakeNIR & DeepfakeNIR & \textbf{100} & \textbf{100} \\
    
    \hline
    CNNDetection  & DeepfakeNIR++ & DeepfakeNIR++ & 89.55 & 96.20 \\
    GANFingerprint  & DeepfakeNIR++ & DeepfakeNIR++ & 64.55 & 72.13 \\
    FCAN-DCT + ResNet50 & DeepfakeNIR++ & DeepfakeNIR++ & \textbf{91.40} & \textbf{98.11} \\
    FCAN-DCT + Xception & DeepfakeNIR++ & DeepfakeNIR++ & 90.96 & 97.58 \\

    \bottomrule
  \end{tabular}
  \label{tab:test_in_ForgeryNIR-V}
\end{table*}

\begin{table}[t]
\centering
  \caption{Cross-dataset evaluation on Celeb-DF (v2) (AUC) via training on FF++ (HQ), where the best result is marked in bold and the next best result is underlined.}
  \begin{small}
  
  \begin{tabular}{ccc}
    \toprule
    Method & FF++ (\%) & Celeb-DF (v2) (\%)\\
    \midrule
     MesoNet  & 83.00 & 53.60 \\
     Xception   & \textbf{99.70} & 65.30\\
     Capsule  & 90.61 & 67.92  \\
     Inproved Xception  & 98.09  & 68.39           \\
     \hline
     FCAN-DCT + ResNet50 & 98.85   & \underline{79.95}           \\
     FCAN-DCT + Xception  & \underline{99.03}  & \textbf{83.46}       \\
    \bottomrule
  \end{tabular}
    \end{small}
  \label{tab:cross-dataset on FF++}
\end{table}

\begin{table}[t]
\centering
  \caption{Cross-dataset evaluation on Celeb-DF (v2) (ACC) via training on WildDF(WildDeepfake), where the best result is marked in bold.}
  \begin{small}
  \begin{tabular}{ccc}
    \toprule
    Method & WildDF (\%) & Celeb-DF (v2) (\%)\\
    \midrule
     MesoNet  & 73.95 & 49.11 \\
     Xception   & 79.99 & 51.87\\
     Capsule  & 78.68 & 53.00  \\
     DAM  & 83.32  & 72.62           \\
     \hline
     FCAN-DCT + ResNet50 & 86.23   & 81.45            \\
     FCAN-DCT + Xception  & \textbf{86.35}  & \textbf{85.74}     \\
    \bottomrule
  \end{tabular}
  \end{small}
  \label{tab:cross dataset on WildDeepfake}
\end{table}

{\bfseries Evaluation on Celeb-DF (v2)}: Considering video-based face forgery detection tends to realize the inconsistency between frames, it seems more reasonable to use 3D-CNN directly considering the temporal relationship, we test some of the most popular networks utilizing temporal features including \textbf{RCN} \cite{41-guera2018deepfake}, \textbf{R2Plus1D} \cite{42-tran2018closer}, \textbf{I3D} \cite{43-carreira2017quo}, \textbf{MC3} \cite{42-tran2018closer} and \textbf{R3D} \cite{44-hara2017learning} on Celeb-DF (v2) \cite{31-li2020celeb}.

Experimental results are shown in Table \ref{tab:test_in_celeb-df}. It can be seen that FCAN-DCT outperforms all existing temporal-based networks to achieve competitive accuracy and AUC with state-of-the-arts. Especially when ResNet50 is selected as the backbone network, the accuracy and AUC reached 99.8\% and 99.99\% respectively which is a near-ideal result. R3D \cite{44-hara2017learning} is a 3D CNNs based on ResNet toward a better action representation. R2Plus1D \cite{42-tran2018closer} decomposes 3D convolutional filters into separate spatial and temporal components and improves the accuracy of action detection. However, few of these methods are specifically designed for video forgery detection and too many parameters of 3D-CNN models make it difficult to apply to lightweight scenarios. Our method not only notices the most important part in different frames but also explores temporal information in the frequency domain and finally presents a comprehensive spatial-temporal  representation. In addition, apart from the backbone network, no additional parameters are added. Therefore, our method is more capable and effective in capturing spatial-temporal forgery inconsistencies than 3D-CNN methods.

\subsection{Cross-dataset experiments}
In this subsection, we conduct a series of cross-dataset experiments to verify the generalization ability of our proposed method. First, we train a classifier based on the high-quality version of FF++ and WildDeepfake respectively, and evaluate on the Celeb-DF(v2). We compare FCAN-DCT with several recently state-of-the-art methods, including \textbf{MesoNet} \cite{32-afchar2018mesonet}, \textbf{Xception} \cite{34-chollet2017xception}, \textbf{Capsule} \cite{35-nguyen2019capsule}, and \textbf{DAM} \cite{39-lin2021improved}.

The comparative results are shown in Table \ref{tab:cross-dataset on FF++} and Table \ref{tab:cross dataset on WildDeepfake} respectively. When trained on FF++ and tested on Celeb-DF (v2), our method achieves AUC metrics of 79.95\% and 83.46\% on ResNet50 and Xception backbone, respectively, outperforming other methods by at least 10\%. In addition, When trained on WildDeepfake, our method also achieves the highest accuracy of 85.74\% on Xception backbone. Our model leverages both spatial and temporal information in frequency domain and is more robust to the modality differences between VIS and NIR because it is in the frequency domain rather than the spatial domain, therefore generalizing better from one method to another which confirms the effectiveness and robustness of our proposed FCAN-DCT.

\subsection{Experiment of Video Forgery Detection in NIR Scenario}
In this subsection, we conducted the experiments with state-of-the-art (SOTA) Deepfake detection methods such as \textbf{CNNDetection} \cite{19-wang2020cnn}, \textbf{GANFingerprint} \cite{17-yu2019attributing} and our proposed FCAN-DCT on our newly constructed DeepfakeNIR and DeepfakeNIR++ dataset, to evaluate the effectiveness and robustness of the proposed method. The experimental results are shown in Table  \ref{tab:test_in_ForgeryNIR-V}. It can be seen that with the help of our method, the model trained only using sources based on the VIS modality achieves over 90\% accuracy in the NIR scenario. Especially for the deeper structure network: Xception \cite{34-chollet2017xception}, when the training set is selected as Celeb-DF (v2) \cite{31-li2020celeb}, the accuracy and AUC reach 98.32\% and 99.90\% respectively. Furthermore, our method achieves 100\% accuracy and 100\% AUC on both ResNet50 and Xception backbone networks when trained on the NIR dataset. On the more challenging DeepfakeNIR++ dataset, our method achieves 91.40\% accuracy and 98.11\% AUC on the ResNet50 backbone, respectively. In particular, whether it is cross-modality experiments on WildDeepfake, Celeb-DF (v2) or experiments on DeepfakeNIR and DeepfakeNIR++, our method achieved the state-of-the-art (SOTA) performance. The considerable performance demonstrates the robustness of our spatial-temporal frequency clue for forgery detection in the VIS and NIR scenario, which can facilitate research on heterogeneous face detection.

\subsection{Ablation Study}
In this section, we conduct sufficient experiments to demonstrate the effectiveness of our method and analyze the impact of hyperparameter selection on the results. In addition, we perform ablation study on the proposed FCAN-DCT. We train and test CFE and FCAN-DCT separately, and the experimental result of each model is shown in Table \ref{tab:ablation_study}. It should be noticed that the temporal clue extracted in FTA is a weight attention map for the CFE module which is designed for extracting the spatial frequency feature. The final spatial-temporal representation can only be obtained with the joint cooperation of both two branches. Therefore, we just operate the CFE module in this section while it is not available to experiment only with the attention weights obtained by the FTA branch.

\begin{table}[t]
\centering
  \caption{Ablation study performed on WildDeepfake dataset. The backbone applied in the network is ResNet50 and the setting of $K=4\times4$, $N=16$.}
  \scalebox{0.8}{
  \begin{tabular}{cccc}
    \toprule
    Weight matrix & Temporal attention map  & Accuracy (\%) & AUC (\%)\\
    \midrule
           -              &                -                    & 83.88 & 92.45 \\
        \checkmark              &                -               & 85.13 & 92.45 \\
        -             &                          \checkmark                  & 85.37 & 93.47 \\
        \checkmark            &                  \checkmark                   & 86.35 & 93.74 \\
    \bottomrule
  \end{tabular}
  }
  \label{tab:ablation_study}
\end{table}



{\bfseries Effectiveness of the cofficient weight matrix}: Previous works \cite{18-frank2020leveraging, 19-wang2020cnn, 20-dzanic2020fourier, 21-chandrasegaran2021closer, 24-li2021frequency} on forgery detection in the frequency domain for a single image believe that there exists a significant difference between the real and fake images in the midium and high frequencies, and this difference is more obvious as the frequency increases. Inspired by these works, we construct a coefficient weight matrix whose height and width are the same as those of the DCT spectrum. While keeping the low-frequency components fixed, we increase the weights of the midium and high-frequency components so that the model pays more attention to the area for fake clues. The experimental results are shown in Table \ref{tab:ablation_study}. When there is no cofficient weight matrix, the accuracy of merely applying the original DCT spectrum reaches 85.37\% while adding the cofficient weight matrix in the frequency transform will increase the accuracy and AUC by 0.95\% and 0.27\%, respectively. The progress demonstrates the cofficient weight matrix is efficient and effective.

{\bfseries Effectiveness of the temporal attention map}: By comparing with and without the temporal attention map settings in FCAN-DCT, it is clear that the results of FCAN-DCT outperform the CFE branch with only the spatial compact feature. For example, when there is no temporal frequency attention map, the accuracy of merely applying the spatial frequency forgery clue reaches 85.13\%. Furthermore, adding the temporal attention map to the spatial clue will increase the accuracy and AUC by 1.22\% and 1.29\%, respectively, which denotes that the temporal attention map learned by FTA is effective in video forgery detection.

{\bfseries Effectiveness of the maximum block strategy}: In the CFE module, we use the maximum strategy to compress the obtained spectrm sequence. Due to the discreteness of DCT, the effective information embedded in DCT frequency spectrum is sparse. To make the frequency feature expression more compact, we first divide the whole spectrum $F$ into $K$ frequency band, and then use the max response to represent each corresponding frequency band. We conduct the strategy comparison experiment by taking the maximum, minimum, and average value within the block on the WildDeepfake datase. Our chosen framework backbone is ResNet50. The experimental results proved that the maximum strategy outperforms minimum as well as average strategy in both effectiveness and efficiency.

\begin{table}[t]
\centering
  \caption{Maximum block strategy ablation study on WildDeepfake dataset, where the best result is marked in bold.}
  \setlength{\tabcolsep}{5.5mm}{
  \begin{tabular}{ccc}
    \toprule
    Method & Accuracy (\%) & AUC (\%)\\
    \midrule
     Min & 83.75 & 91.85 \\
     Avg  & 83.38 & 91.14\\
     Max (Ours) & \textbf{86.35} & \textbf{93.74}\\
    \bottomrule
  \end{tabular}
  }
  \label{tab:test_in_celeb-df}
\end{table}

\section{CONCLUSIONS}

In this paper, we present a novel forgery detection method based on spatial-temporal frequency clue. We then performed comprehensive experiments on two widely used VIS video forgery datasets and our newly constructed NIR video forgery dataset DeepfakeNIR. The experimental results show that our method outperforms existing methods on both accuracy and AUC, and has excellent robustness and generalization ability in heterogeneous video forgery detection scenarios. Furthermore, the proposed novel dataset DeepfakeNIR based on the NIR modality will further facilitate future research on forgery detection in the near-infrared modality. In the future, we intend to explore the generalizability of video forgery detection in the presence of adding more real-world perturbations as well as adversarial perturbations. 

\bibliography{aaai22}

\end{document}